\documentclass[]{IOS-Book-Article}

\usepackage{nesybook}

\newcommand{\citep}[1]{\cite{#1}}

\usepackage{amsmath}
\DeclareMathOperator*{\argmax}{arg\,max}

\usepackage{chapter-graph-querying/chap}

\begin{document}

% \pagestyle{headings}
% \def\thepage{}

% XXX

% \markboth{May 2020\hb}{May 2020\hb}
%\maketitle

%\setcounter{tocdepth}{0}
%\tableofcontents

%\setcounter{page}{0}
% \pagestyle{headings}

%\include{chapter-graph-querying/chap}

% !TeX root = ../nesybook.tex

\chapter{Approximate Answering of Graph Queries}
\label{chapter-approx-graph-queries}
%\item 
%\item 
%\item 
%\item 
%\item 
%\item 
%\item 
%\item 

\chapterauthor{Michael Cochez}{Vrije Universiteit Amsterdam \& Elsevier Discovery Lab}
\chapterauthor{Dimitrios Alivanistos}{Vrije Universiteit Amsterdam \& Elsevier Discovery Lab}
\chapterauthor{Erik Arakelyan}{University of Copenhagen }
\chapterauthor{Max Berrendorf}{Ludwig-Maximilians-Universität München}
\chapterauthor{Daniel Daza}{Vrije Universiteit Amsterdam \& Elsevier Discovery Lab}
\chapterauthor{Mikhail Galkin}{Mila Quebec \& McGill University}
\chapterauthor{Pasquale Minervini}{University of Edinburgh}
\chapterauthor{Mathias Niepert}{University of Stuttgart}
\chapterauthor{Hongyu Ren}{Stanford University}

\allchapterauthors{
Michael Cochez,
Dimitrios Alivanistos,
Erik Arakelyan,
Max Berrendorf,
Daniel Daza,
Mikhail Galkin,
Pasquale Minervini,
Mathias Niepert,
Hongyu Ren}

Knowledge graphs (KGs) are inherently incomplete because of incomplete world knowledge and bias in what is the input to the KG.
Additionally, world knowledge constantly expands and evolves, making existing facts deprecated or introducing new ones.
However, we would still want to be able to answer queries \emph{as if the graph were complete}.
In this chapter, we will give an overview of several methods which have been proposed to answer queries in such a setting. 
We will first provide an overview of the different query types which can be supported by these methods and datasets typically used for evaluation, as well as an insight into their limitations.
Then, we give an overview of the different approaches and describe them in terms of expressiveness, supported graph types, and inference capabilities.

\section{Introduction}

Knowledge Graphs are used to represent complex information from multiple domains.
This knowledge is represented using a graph consisting of nodes and directed labeled edges.
An edge connecting two nodes indicates that the source node has a relation of a specific label with the second node.
Nodes in a knowledge graph are often called entities. 
They can refer to real-world entities or be literal values, such as the foundation date of a city or the name of a person.

Systems built around knowledge graphs can leverage the information in them through the use of \emph{graph queries}.
With these, a query answering system can find specific entities or relations that fulfill the constraints specified in the query.
Besides, one can perform manipulations over the retrieved results, such as aggregations, similar to what is possible in a relational database.
A recent overview covering these and many other aspects of knowledge graphs can be found in the survey by Hogan et al.~\cite{hogan_knowledge_2021}.

Due to the incomplete nature of knowledge graphs, a query answering system might not return a complete set of answers because the graph does not contain all information needed to perfectly match the conditions in a query.
To address this setting, \emph{approximate query answering} methods have been proposed, which attempt to provide likely answers to a query by 
learning from patterns in the graph and generalizing to queries that rely on missing edges.

In the remainder of this chapter, we will first give a formal definition of the approximate query answering task and an overview of datasets and metrics used for performance evaluation.
Furthermore, we will introduce different methods proposed in the literature, where we note that the order in which they are presented is not in historical order of appearance but based on type.
While this chapter provides an overview of several methods used for this task, it does not aim for completeness.
We refer to \cite{ren2023neural} for a comprehensive overview of different methods.

\section{Definitions}
\label{sec:definitions}
%\todo[inline]{MC Here we write our problem definition}
%
%\todo[inline]{MC Assigned to Mikhail Galkin}
%
%\todo[inline]{Query Complexity}

%\begin{itemize}
%	\item Knowledge Graph
%	\item Query (not hyper-relational) - 
%	\item multi-hop reasoning
%	\item 
%\end{itemize}

To perform the approximate query answering, we need a knowledge graph (KG). 
The most commonly used graph type in this domain is the heterogeneous edge labeled graph; the definition below is slightly adapted from~\cite{hogan_knowledge_2021}.

\begin{kgqadefinition}[Knowledge Graph]
Given a countably infinite set of constants $\mathbf{Con}$.
A \emph{Knowledge Graph} is a tuple $G = (V,E,L,l)$, where $V \subseteq \mathbf{Con}$ is a set of nodes, $L \subseteq \mathbf{Con}$ is a set of edge and node labels, $E \subseteq V \times L \times V$ is a set of edges, and $l : V \rightarrow L$ maps each node to a label.
\noindent Nodes and edge labels cannot be present without any associated edge.
\end{kgqadefinition}

\noindent Next, we give our definition of a query.
The basis for our definition in this chapter relies on the definition of directed edge-labelled graph patterns from \cite{hogan_knowledge_2021}. We first define a countably infinite set of \textit{variables} $\mathbf{var}$ ranging over (but disjoint from: $\mathbf{Con} \cap \mathbf{Var} = \emptyset$) the set of constants. 
The union of the constants and the variables are called the \textit{terms}, $\mathbf{Term} = \mathbf{Con}\cup \mathbf{Var}$.

\begin{kgqadefinition}[Query]
We define a \emph{Query} as a tuple $Q = (V',E',L')$, where $V' \subseteq \mathbf{Term}$ is a set of node terms, $L' \subseteq \mathbf{Term}$ is a set of edge terms, and $E' \subseteq V' \times L' \times V'$ is a set of edges (triple patterns).
Again, we require that no node terms or edge terms can be present without an associated edge.
\end{kgqadefinition}

Finding an answer to such a query is equivalent to finding a partial mapping $\mu : \mathbf{Var} \to \mathbf{Con}$, such that if each variable $v$ in the query gets replaced by $\mu(v)$, then the resulting triples are part of the edge set of the Knowledge Graph.
To get all answers to the query, we need to find all possible partial mappings.

\noindent Most techniques for approximate query answering only deal with a limited set of query types.
Usually, the supported types are a subset of the queries defined in this chapter.
In particular, hardly any of the existing methods allows for the edge terms to be a variable. 
Aside from that, many of the methods require that all queries have very specific structures.
See \cref{subsec:training-regimes} for more details. 

Furthermore, approximate query answering techniques are often evaluated using queries where there is only one variable of interest, which is called the \emph{target variable}. 

We further note that some approaches are more naturally explained with an equivalent formalism where triples are expressed as binary predicates, i.e., each triple of the form $(s,p,o)$, can be written in first-order logic as $p(s,o)$, meaning that the relation $p$ holds between $s$ and $o$, with $p$ contained in a set of relation labels $R\subseteq L$ and $s,p\in E$.
Similarly, the query can be formulated using existentially quantified variables, as a conjunction of formulas where the binary predicates are applied to constants and variables.
In this formalism, one can also define further operations if one also introduces negation of a formula and disjunction (see, for example, BetaE in \cref{sec:betaE}).

Finally, we can define what we mean by Approximate Query Answering, which we base on the definition in \cite{alivanistos2022hyper}.
Given an incomplete Knowledge Graph $G = (V,E,L,l)$ (part of the not observable complete KG $\hat{G}$) and a query Q, with one of its variables $t$ chosen as the target variable of the query, the complete set of answers $T$ to this query is the set of all values for the mappings of $t$. 
$$T = \lbrace \mu(t) | \mu \text{ is an answer to the query when executed \linebreak[0] over the complete KG } \hat{G} \rbrace$$
The task of \emph{Approximate Query Answering} is to rank all nodes in $V$ \footnote{If variables are allowed as edge terms, one would rank these.} such that the  elements of $T$, are at the top of the ranking.
Since the given KG is incomplete, we cannot solve this problem directly as a graph-matching problem, as usually done in databases.

\section{Evaluating Approximate Query Answering Performance}

To evaluate the different methods designed for approximate query answering, various datasets and evaluation metrics have been proposed.
We present both in this section. 
The reader should be warned that not all approaches have been evaluated with all these datasets and metrics. Some of them are just not applicable, while others have fallen out of favor.
See also the overview in \cref{tab:query:datasets} in the appendix.

\subsection{Datasets}

Generally, datasets for approximate graph query answering comprise an incomplete knowledge graph and pairs of test queries with their respective set of answer entities.
Since the methods are developed to answer queries over \emph{incomplete} knowledge graphs, these queries have some answers which cannot be obtained by a simple graph traversal but require implicit knowledge graph completion or reasoning.
These answers are often called the \emph{hard} answers, in contrast with the \emph{easy} answers for this query, which can be found using graph traversal.

The employed datasets are usually constructed with the following process:
Starting from a source knowledge graph, e.g., subsets of Freebase~\cite{bollacker2008freebase}, NELL~\cite{carlson2010toward}, or Wikidata~\cite{vrandevcic2014wikidata}, a train-test split is performed (sometimes known splits from the link prediction literature is reused e.g., FB15k-237~\cite{toutanova2015representing} or NELL-995~\cite{xiong2017deeppath}).
As a result, we have two sets of edges: the train edges form the incomplete knowledge graph, while the test edges are kept out.

Next, we create test queries using several query patterns which yield small subgraphs.
This pattern is matched on the \emph{original} graph before splitting. 
However, at least one edge of this match must be a test edge.
This way, we ensure that in order to answer the query, the methods must deal with incompleteness.
From this matching step, we record the set of answer entities.

\begin{figure}
	\centering
	\includegraphics[width=0.7\linewidth]{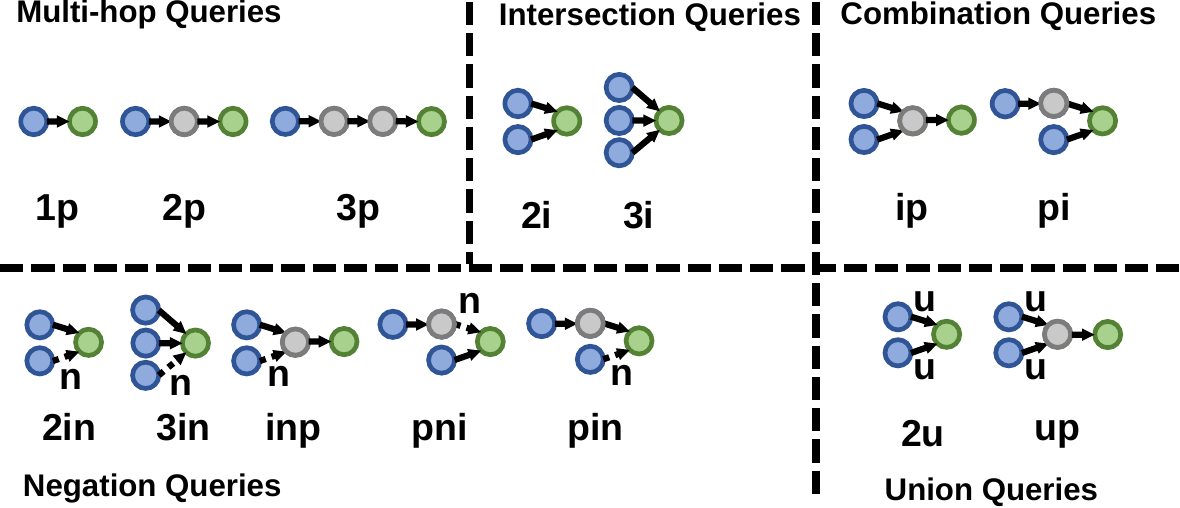}
	\caption{
		Different query patterns are used for the evaluation of approximate query answering systems.
		In these queries, a blue node indicates a constant, in this context also called an anchor.
		The gray nodes indicate variables, meaning that during query execution these can be matched with any node in the KG.
		A green node indicates the target of the query, a special variable; the set of nodes that can be matched to this node is the answer to the query.
		In this diagram, the \texttt{p} indicated projection, \texttt{i} is the intersection, \texttt{u} stands for union and \texttt{n} stands for negation.
	}
	\label{fig:queries}
\end{figure}

Existing methods have been evaluated with different sets of query patterns. 
An overview of existing patterns (adapted from \cite{ren2020beta})can be found in \cref{fig:queries}.

\begin{itemize}[-]
\item On the top left, we see multi-hop queries, also called projection queries because of how they are solved in some of the methods.
These queries start with a specific node, which we call the anchor node, then there are one or more hops in the query. At the end of this chain, we find the target node.
% \todo[inline]{MC We could include a formal definition for each if we want, but this seems overkill.}
\item On the top middle, there are the intersection queries. 
These start from multiple anchor nodes, and each of these is connected to the same target node.
\item On the right, there are combinations of having multiple hops and intersections at the same time.

\item At the bottom we have query patterns for which additional operators are needed.
On the left, we find patterns that have a form of negation, namely that the edge indicated with an \texttt{n} must not exist between the entities.

\item On the right, we see queries that involve a union operator. 
The set of entities which is an answer to the target of the edges marked with \texttt{u}, is the union of the answers when for each of these edges a 1-hop query was answered.

\end{itemize}

\noindent For specific methods, there could be additional specialized patterns. StarQE \cite{alivanistos2022hyper}, for example, is made to answer queries with additional context information and uses patterns that include this. 

\subsection{Training Regimes}
\label{subsec:training-regimes}
When designing an approximate query answering method, one is free to use the training triples as desired. 
Some methods, such as the ones in section \ref{sec:pre-trained}, will use this graph to compute graph embeddings which are then subsequently used to answer a query.
Many other methods will use this graph to create training queries that can be used to train a machine learning model.
This works by first separating a held-out set from the training triples, which we will call the validation triples.
Training queries are created similarly to how the test queries were constructed above: we use the query patterns and guarantee that there is always one edge in the matched pattern that comes from the held-out set.
The idea is that if we are able to train a model to answer these  queries, then it is likely that we can use that to also answer the queries from the test set.

For all approaches, there is one additional aspect of interest, namely that of generalization.
The question here is whether methods that have only seen queries with certain patterns can generalize to unseen patterns. 
If this is the case, then we call the method \emph{generalizable}.

\subsection{Metrics}

During the evaluation, query answering is posed as a ranking task where scores are assigned to candidate entities (i.e., potential answers), with larger scores indicating preferred answers.
Metrics are computed based on this ranking and the complete set of answer entities.
It is worth noting that due to the construction of the datasets and the open-world nature of knowledge graphs, where missing links do not necessarily indicate that the fact is wrong, these complete answer sets may be subsets of the true answer set\footnote{The true answer set is the answer set which contains all answers that would be correct in the real world. This might include answers which cannot even be found in the original graph simply because that one was incomplete to start with. Besides, it is also possible that the original graph has erroneous links. In that case, there is no guarantee that the complete set is a subset of the true set.}.

In general, there are two types of evaluation metrics:
The first uses the area-under-the-curve of the receiver-operator or precision-recall curve, where the former leads to AUROC score while the latter is called APS~\cite{manning2008introduction}. These can only be computed for methods that do not produce a ranking but rather produce a set of answers to the query.

The other type of metric is rank-based.
Here, we determine the \emph{rank} of complete answers, determined by their position when sorting all candidates by score in descending order.
Lower ranks (higher scores) for entities that are answers to the query generally indicate better performance.
Since there may be multiple correct answers to one query, the \emph{filtered} evaluation protocol~\cite{DBLP:conf/nips/BordesUGWY13} is commonly used, where other entities from the correct answer set are ignored when calculating the rank for a given entity.
Each combination of a query and an answer entity yields a single rank.
To obtain a single-figure measure of performance, two metrics are commonly used:
The \emph{mean reciprocal rank (MRR)} is the arithmetic mean of reciprocal ranks, i.e., the inverse of the harmonic mean of individual ranks~\cite{hoyt2022unified}.
The \emph{hits at $k$ (H@k)} metric is the relative frequency of ranks being at most $k$, i.e., of the correct answer entity occurring under the first $k$ entries.

Some works have used weighted rank-based metrics~\cite{ren2020query2box,alivanistos2022hyper}, where they first aggregate rank-based metrics per query and then average across queries.
This ensures that the performance on each query contributes the same amount to the total score, irrespective of the number of answer entities in those queries.
Notice that in the presence of exactly equal scores, the ordering may be ambiguous and the tie-break policy can have a considerable impact on the ranks~\cite{berrendorf2020interpretableArxiv}.

Finally, we want to note that newer works often omit precision and recall, while at least one work \cite{van_bakel_approximate_2021} calls for additional evaluation of these methods with precision and recall metrics.
This both has a benefit and a drawback.
The benefit is that methods have to create a crisp set of answers, which is important if these systems are used in applications that cannot deal with rankings. 
Similarly, if these systems are used as part of a more feature-rich query execution engine, then many operations are only possible if the sub-queries result in sets.
A major issue with this type of evaluation is, however, that the original graphs used for creating the evaluation sets are themselves not complete, which results in less favorable results in these metrics, even if the answer is correct. 
We note that in link prediction literature earlier works were also using precision-recall, while later ones resorted to ranking-based metrics.

\begin{table}
	\centering
	\caption{Overview of used metrics in the investigated papers. * denotes that metrics are first aggregated per query and then averaged across queries. This leads to more balanced weights in the presence of queries with vastly different numbers of answers.}
	\label{tab:metrics}
	\begin{tabular}{lllll}
		\toprule
		&
		\rotatebox{90}{AUC} &
		\rotatebox{90}{APR} &
		\rotatebox{90}{H@k} &
		\rotatebox{90}{MRR}
		\\
		\midrule
		AQRE~\cite{wang2018towards} & ? \\
		GQE~\cite{hamilton2018embedding} & \checkmark & \checkmark & - & - \\
		MPQE~\cite{daza2020message} & \checkmark & \checkmark & - & - \\
		BiQE~\cite{biqe} & \checkmark & \checkmark & \checkmark & \checkmark \\
		BetaE~\cite{ren2020beta} & - & - & \checkmark & \checkmark \\
		Q2B~\cite{ren2020query2box} & - & - & \checkmark* & \checkmark*\\
		CQD~\cite{arakelyan2021complex} & - & - & \checkmark & - \\
		StarQE~\cite{alivanistos2022hyper} & - & - & \checkmark* & \checkmark* \\
		\bottomrule
	\end{tabular}
\end{table}

\section{Approaches with Pre-trained Graph Embeddings}
\label{sec:pre-trained}

%\subsection{AQRE}
%\cite{wang2018towards}
%
%\todo[inline]{Assigned to Wang Meng}

\subsection{Continuous Query Decomposition}
% \cite{arakelyan2021complex}
% \todo[inline]{Assigned to Erik Arakelyan/Pasquale Minervini}

%
Continuous Query Decomposition~\cite{arakelyan2021complex} (CQD) is a framework for answering First-Order Logic Queries, where the query is compiled in an end-to-end differentiable function, modeling the interactions between its atoms (corresponding to triples, or binary predicates).
The truth value of each atom is computed by a neural link predictor~\citep{nickel2015review} that, given an atomic query, returns the likelihood that the fact it represents holds true.
Then, the authors propose two approaches for identifying the most likely values for the variable nodes in a query -- either by continuous or by combinatorial optimization.
More formally, given a query $Q$, we define the score of an entity $a \in E$ as a candidate answer for a query as a function of the score of all atomic queries in $Q$, given a variable-to-entity substitution for all variables in $Q$.
Each variable is mapped to an \emph{embedding vector}, that can either correspond to an entity $c \in E$ or to a virtual entity.
The score of each of the query atoms is determined individually using a neural link predictor~\citep{nickel2015review}.
Then, the score of the query with respect to a given candidate answer $Q[a]$ is computed by aggregating all of the atom scores using t-norms and t-conorms -- continuous relaxations of the logical conjunction and disjunction operators.
\paragraph{Atomic Query Answering via Neural Link Predictors}
A neural link predictor is a differentiable model where atom arguments are first mapped into a $k$-dimensional embedding space and then used for producing a score for the atom.
More formally, given a query atom $p(s, o)$, where $p \in R$  and $s, o \in E$, the score for $p(s, o)$ is computed as $\phi_{p}(\mathbf{e}_{s}, \mathbf{e}_{o})$, where $\mathbf{e}_{s}, \mathbf{e}_{o} \in \mathbb{R}^{k}$ are the embedding vectors of $s$ and $o$, and $\phi_{p} : \mathbb{R}^{k} \times \mathbb{R}^{k} \mapsto [0, 1]$ is a \emph{scoring function} computing the likelihood that entities $s$ and $o$ are related by the relationship $p$.
In CQD~\cite{arakelyan2021complex}, the authors use a regularized variant of ComplEx~\citep{trouillon2016,lacroix2018canonical}, but it is possible to replace it with an arbitrary neural link predictor.
\paragraph{T-Norms} 
A \emph{t-norm} $\top : [0, 1] \times [0, 1] \mapsto [0, 1]$ is a generalization of conjunction in logic~\citep{DBLP:books/sp/KlementMP00,DBLP:journals/fss/KlementMP04a}.
Some examples include the \emph{Gödel t-norm} $\top_{\text{min}}(x, y) = \min\{ x, y \}$, the \emph{product t-norm} $\top_{\text{prod}}(x, y) = x \cdot y$, and the \emph{Łukasiewicz t-norm} $\top_{\text{Luk}}(x, y) = \max \{ 0, x + y - 1 \}$.
Analogously, \emph{t-conorms} are dual to t-norms for disjunctions -- given a t-norm $\top$, the complementary t-conorm is defined by $\bot(x, y) = 1 - \top(1 - x, 1 - y)$.
\paragraph{Continuous Reformulation of Complex Queries}
Let $Q$ denote the following Disjunctive Normal Form (DNF) query:
\begin{align}
& Q[A] \triangleq ?A : \exists V_{1}, \ldots, V_{m}.\left( e^{1}_{1} \land \ldots \land e^{1}_{n_{1}} \right) \lor \ldots \lor \left( e^{d}_{1} \land \ldots \land e^{d}_{n_{d}} \right), \label{eq:opt} \\
& \quad \text{where} \; e^{j}_{i} = p(c, V), \ \text{with} \; V \in \{ A, V_{1}, \ldots, V_{m} \}, c \in E, p \in R, \notag \\
& \quad \text{or} \; e^{j}_{i} = p(V, V^{\prime}), \ \text{with} \; V, V^{\prime} \in \{ A, V_{1}, \ldots, V_{m} \}, V \neq V^{\prime}, p \in R. \notag
\end{align}
CQD tries to identify the variable assignments that render the query $R$ in \cref{eq:opt} true.
To achieve this, CQD casts this as an optimization problem, where the aim is finding a mapping from variables to entities that \emph{maximizes} the score of $Q$:
\begin{align}
& \argmax_{A, V_{1}, \ldots, V_{m} \in E} \left( e^{1}_{1} 
\top \ldots \top e^{1}_{n_{1}} \right) \bot \ldots \bot \left( e^{d}_{1} \top \ldots \top e^{d}_{n_{d}} \right) \label{eq:cont} \\
& \quad \text{where} \; e^{j}_{i} = \phi_{p}(\mathbf{e}_{c}, \mathbf{e}_{V}), \ \text{with} \; V \in \{ A, V_{1}, \ldots, V_{m} \}, c \in E, p \in R \notag \\
& \quad \text{or} \; e^{j}_{i} = \phi_{p}(\mathbf{e}_{V}, \mathbf{e}_{V^{\prime}}), \ \text{with} \; V, V^{\prime} \in \{ A, V_{1}, \ldots, V_{m} \}, V \neq V^{\prime}, p \in R, \notag
\end{align}
\noindent where $\top$ and $\bot$ denote a t-norm and a t-conorm -- a continuous generalization of the logical conjunction and disjunction, respectively -- and $\phi_{p}(\mathbf{e}_{s}, \mathbf{e}_{o}) \in [0, 1]$ denotes the neural link prediction score for the atom $p(s, o)$.
Note that, in \cref{eq:cont}, the bound variable nodes $V_{1}, \ldots, V_{m}$ are only used through their embedding vector: to compute $\phi_{p}(\mathbf{e}_{c}, \mathbf{e}_{V})$ CQD only uses the embedding representation $\mathbf{e}_{V} \in \mathbb{R}^{k}$ of $V$, and does not need to know which entity the variable $V$ corresponds to.
For this reason, in Arakelyan et al. \cite{arakelyan2021complex}, the authors propose two possible strategies for finding the optimal variable embeddings $\mathbf{e}_{V} \in \mathbb{R}^{k}$ with $V \in \{ A, V_{1}, \ldots, V_{m} \}$ for maximizing the objective in \cref{eq:cont} in CQD, namely \emph{continuous optimization}, where they optimize $\mathbf{e}_{V}$ using gradient-based optimization methods, and \emph{combinatorial optimization}, where they search for the optimal variable-to-entity assignment.

\paragraph{Complex Query answering via Continuous and Combinatorial Optimization}
One way the authors solve the optimization problem in \cref{eq:cont} is by finding the variable embeddings that maximize the score of a complex query.
\cref{eq:cont} is modified into the following continuous optimization problem:
%
%
%\begin{equation}
\begin{align}
\argmax_{\mathbf{e}_{A}, \mathbf{e}_{V_{1}}, \ldots, \mathbf{e}_{V_{m}} \in \mathbb{R}^{k}} & \left( e^{1}_{1} 
\ \top \ \ldots \ \top \ e^{1}_{n_{1}} \right) \ \bot \ .. \ \bot \ \left( e^{d}_{1} \ \top \ \ldots \ \top \ e^{d}_{n_{d}} \right) \label{eq:co}
\end{align}

The authors opt to directly optimize the embedding representations $\mathbf{e}_{A}, \mathbf{e}_{V_{1}}, \ldots, \mathbf{e}_{V_{m}} \in \mathbb{R}^{k}$ of variables $A, V_{1}, \ldots, V_{m}$, rather than exploring the combinatorial space of variable-to-entity mappings.
In this way, it is possible to tackle the maximization problem in \cref{eq:co} using gradient-based optimization methods, such as Adam. %~\citep{kingma2014adam}.
After identifying the optimal representation for variables $A, V_{1}, \ldots, V_{m}$, the method replaces the query target embedding $\mathbf{e}_{A}$ with the embedding representations $\mathbf{e}_{c} \in \mathbb{R}^{k}$ of all entities $c \in E$. It uses the resulting complex query score to compute the likelihood that such entities answer the query.
This variant of CQD is denoted as CQD-CO.

The combinatorial way we tackle the optimization problem in \cref{eq:cont} is by iteratively searching for a set of variable substitutions $S = \{ A \leftarrow a, V_{1} \leftarrow v_{1}, \ldots, V_{m} \leftarrow v_{m} \}$, with $a, v_{1}, \ldots, v_{m} \in \mathcal{E}$, that maximizes the complex query score, in a procedure akin to \emph{beam search}, thus denoting the method as CQD-Beam.

In CQD-Beam, we traverse the query graph and, whenever we find an atom in the form $p(c, V)$, where $p \in R$, $c$ is either an entity or a variable for which we already have a substitution, and $V$ is a variable for which we do not have a substitution yet, we replace $V$ with all entities in $E$ and retain the top-$k$ entities $t \in E$ that maximize $\phi_{p}(\mathbf{e}_{c}, \mathbf{e}_{t})$ -- the most likely entities to appear as a substitution of $V$ according to the neural link predictor.

\section{Projection Approaches}
\subsection{GQE}

Graph Query Embedding (GQE)~\cite{hamilton2018embedding} was introduced as one of the first methods for approximate query answering over KGs. It is focused on conjunctive queries that can be represented as a Directed Acyclic Graph (DAG), which contains nodes for entities known in the query (called anchor nodes), variables, and a target node.

GQE operates by learning an embedding for each entity in the graph, which is randomly initialized before training with query-answer pairs.
For a given query, it performs a sequence of operations that follow the structure of its corresponding DAG. The operations start from the embeddings of the entities at the anchor nodes, and progressively apply projection and intersection operators until reaching the target node. 

The projection operator in GQE is employed when traversing a path from one node to another, and the intersection operator is used when two or more paths collide in a node. The operators are defined as functions of the embeddings of entities and relations involved. For a given entity embedding $\mathbf{e}_i$ and relation type $r$,  the projection operator $\mathcal{P}$ and the intersection operator $\mathcal{I}$ are defined as

\begin{align}
\mathcal{P}(\mathbf{e},r) &= \mathbf{R}_r\mathbf{e} \\
\mathcal{I}(\lbrace\mathbf{e}_1, \ldots, \mathbf{e}_n\rbrace) &= \mathbf{W} f(\lbrace\mathbf{e}_1,\ldots, \mathbf{e}_n \rbrace),
\end{align}

where $\mathbf{R}_r$ is a matrix associated with relation $r$, $\mathbf{W}$ is a matrix, and $f$ is a function that maps a variable sequence of embeddings (as many as those colliding into a node) to a single embedding. In GQE, $f$ is implemented as a multi-layer perceptron applied to each embedding, followed by an element-wise mean.

Once the operations reach the target node, the result is an embedding called the \emph{query embedding}. All parameters in GQE (i.e. the entity embeddings, and all parameters in the operators) are optimized such that the query embedding has a high dot product with the embedding of entities that are answers to the query, and a low dot product with embeddings of entities selected at random. As such, GQE requires generating a training dataset of millions of query-answer pairs with sufficient coverage for different query structures. Crucially, the training dataset must contain queries that allow the training of both the intersection and projection operators. This is in contrast with methods like CQD, which is trained for 1-hop link prediction and does not make use of learnable parameters for complex queries.

\subsection{Query2Box}
Query2box~\cite[Q2B,]{ren2020query2box} proposes to embed queries as hyper-rectangles (boxes) with a center embedding and offset embedding. 
Compared with GQE, Q2B provides a better way to represent queries or sets. Similarly, Q2B designs neural relational projection and neural intersection operators. The projection operator simulates the mapping from a fuzzy set to another fuzzy set by translating the boxes and simulates set intersection using box intersection with attention. 
Given an entity embedding (a point vector) and a query embedding (a box), the distance function in Q2B is a modified L1 distance: a weighted sum of the in-box distance and out-box distance, where the in-box distance is down-weighted. Q2B is the first query embedding method that handles any existential positive first-order query with a tree-structured computation graph. The idea is to convert any input query to its equivalent disjunctive normal form (DNF), such that each conjunctive component can be embedded and the overall distance between an entity and a query is the minimum of the distance between the entity and all the conjunctive queries in the DNF.

\subsection{BetaE}
\label{sec:betaE}
BetaE~\cite{ren2020beta} is a query embedding method with a probabilistic embedding space. The insight is to embed entities and queries as multiple independent Beta distributions. The key capability of BetaE is that it can faithfully handle a broader set of logical operators, especially logical negation (or set complement) operators. Handling negation is challenging for previous query embeddings such as GQE and Query2box since the complement of a point or a box in the euclidean space cannot be approximately by a point or a box. Due to the flexibility of Beta distribution, one can easily flip the density by taking the reciprocal of the parameters of a Beta distribution compared with prior query embeddings with a Euclidean embedding space. This simple neural negation operator also satisfies the property that taking negation/complement twice gives an identity mapping. Accordingly, BetaE also designs relation projection and neural intersection operators. For relation projection, BetaE employs an MLP that maps from the Beta embedding of the input set to that of the output. For intersection, the Beta embedding of the intersected set is a weighted product of the input since the product of several Beta distributions is still proportional to a Beta distribution. The neural intersection operator also satisfies several properties, including $\mathcal{I}(S,S)=S$. Besides, BetaE also introduces uncertainty/cardinality modeling, which is a metric to evaluate query embeddings' ability to capture the cardinality of a query with the learned embeddings. Together, BetaE gives rise to the first approximate query-answering models that handle conjunction, disjunction, and negation at the same time.

\section{Message Passing Approaches}

\subsection{MPQE}
% \todo[inline]{Assigned to Daniel Daza/Michael Cochez}
While GQE~\cite{hamilton2018embedding} demonstrated the effectiveness of using embeddings for complex query answering, the presence of specific operators for projections and intersections, and the algorithm for obtaining the query embedding results in a limited set of query structures that can be embedded and a need for a large set of query-answer pairs.

Message Passing Query Embedding~\cite[MPQE,]{daza2020message} set to address these limitations by noting that the projection and intersection operators can be generalized into a message passing operation over the query DAG, for which numerous architectures have been proposed in the research area of graph neural networks (GNNs)~\cite{wu2020comprehensive,zhou2020graph}.

Similar to GQE, MPQE learns embeddings of entities in the KG. Additionally, it also learns generic \emph{type embeddings}, which are used for nodes at the variable positions in the query DAG. The query embedding is computed by first assigning embeddings to all nodes in the DAG: anchor nodes are assigned entity embeddings, the variable and target nodes are assigned embeddings corresponding to the type of the variable (such as \emph{Person}, or \emph{Drug}), and the edges are assigned a relation type embedding. The graph and embeddings are then passed as input to a GNN, which updates the embeddings of nodes in the graph. After an arbitrary number of updates, the node embeddings are reduced to a single query embedding by means of an aggregation such as the mean, sum, or element-wise maximum. The authors also proposed a dynamic aggregation function motivated by the fact that in a query with diameter\footnote{The diameter of a query DAG is defined as the longest shortest path between two nodes in the graph.} $d$, a GNN would require $d$ message-passing steps to guarantee that a node has received messages from all the nodes in the graph. Experiments show that when trained with chain-like queries only, the dynamic form of MPQE is significantly better than other aggregation functions. However, when training over a broader set of query structures, there is no clear favorable function.

Since in principle the GNN in MPQE can encode any graph, it is not limited to DAGs. Besides, it can be trained with chain-like queries, similar to CQD, while still retaining good performance for complex queries at test time. 

\subsection{StarQE}
Approximate query answering methods have also been extended to a type of graph known as \emph{hyper-relational} knowledge graphs.
Hyper-relational KGs allow for qualifying or quantifying relations between nodes. An example can be seen with the relation \textit{educated at} and \textit{Albert Einstein}. \textit{Albert Einstein} was educated at two different institutions, namely \textit{ETH} and \textit{University of Zurich}. In a hyper-relational KG, we can introduce \emph{qualifiers} for facts with fine-grained information, for example, that \textit{Albert Einstein} was educated at \textit{ETH} with a \textit{degree} of BSc.
Under the standard definition of a KG in section~\ref{sec:definitions}, querying for the places where Albert Einstein received his education would retrieve both \textit{ETH} and \textit{University of Zurich} as correct answers. However, the addition of a qualifier \textit{degree}, indicating the degree received during that study, would narrow down the result set of the query to a single correct answer. 

StarQE is a recently proposed method for query answering over hyper-relational KGs based on a message-passing approach~\cite{alivanistos2022hyper}. It relies on StarE~\cite{galkin2020message}, a specialized graph neural network to encode hyper-relational graphs, to obtain an embedding of hyper-relational query in a similar vein as in MPQE.

The authors show that the narrowing down of the result set leads to a substantial increase in performance over different query patterns.
In their work, the authors introduced a new query dataset called \textsc{WD50K-QE} which is generated by querying a hyper-relational subset of Wikidata using various specialized logical query patterns.

\section{Approaches using Sequence Encoders}

\textsc{BiQE}~\cite{biqe} computes approximate answers to conjunctive queries with one or more free variables but limits the class of queries to those representable with a directed acyclic graph (DAG), similar to previously described approaches.

A Transformer-based language model~\cite{DBLP:conf/nips/VaswaniSPUJGKP17}, which operates on sequences of tokens, is used for encoding the query DAG. DAGs differ from sequences in two important aspects. First, nodes in graphs can have multiple predecessors and successors, while tokens in a sequence have only, at most, one predecessor and one successor. 
Second, nodes in DAGs are not totally but only partially ordered. As a consequence, using DAGs as input for Transformer models is not straightforward.

\textsc{BiQE} addresses this challenge by decomposing the query DAG into a finite set of paths from each root node to each leaf node of the query DAG. 
For general query DAGs, the number of paths can be exponential in the number of nodes in the worst case. However, this is not a problem in practice as the size of conjunctive queries is typically small and assumed fixed. Indeed, the \emph{combined} query complexity of conjunctive queries, that is, the complexity with respect to the size of the data and the query itself, is NP-hard in relational databases. \textsc{BiQE} uses positional encodings to represent the order of the resulting query paths. A reconstruction loss on many generated queries with a certain fraction of dropped-out elements is used for training.  

\section{Conclusion, Outlook, and Challenges}

The Query Answering approaches described in this chapter share certain strengths and limitations. 
The first limitation is the resources required when training such models. 
As discussed in \cref{subsec:training-regimes}, many approaches need generated queries.
This gives rise to issues with \emph{scalability}, both in training data generation and evaluation of approaches. 
Currently, the methods can only be trained on a fraction of the total number of queries. 
It might be that there is a clever way of sampling queries such that it would improve performance, but current systems only use random sampling.

Another issue is that real-world KGs as well as the benchmark KGs are characterized by popular nodes that can skew the distribution of queries due to their high in or out degree. 
In the paper by Alivanistos et al.~\cite{alivanistos2022hyper} (appendix C), the authors showed that if the 3i test queries would be generated naively, one could achieve a Hits@10 performance of 0.93 by always predicting the top 10 answers in descending popularity, urging the authors to remove high degree nodes.
At the same time, generating a complete set of queries can easily become intractable, making it also more difficult to study this particular issue. 

% COMPLETENESS OVER THE SET OF POSSIBLE LOGICAL OPERATORS
A further issue with the current state of methods is that they only support a limited set of operators. 
Besides, even when a large set is supported, there would still be queries which cannot be represented with them. A prominent example are queries which have cycles in them. While some of the methods (e.g., the message passing based ones) could in principle compute an embedding for these, there is little theoretical basis supporting the idea, and it has not been evaluated.
It is clear that there is still a large scope for exploration left in this aspect.
Besides this, all current methods require the queries to have at least one anchor node, while in general, queries can also consist of only variables.

% WORKING WITH LITERALS
Furthermore, a common challenge amongst the approaches introduced in this chapter appears when trying to incorporate \textit{literal values} in the queries. 
Examples can be \textit{time}, \textit{strings}, numerical values indicating pressure or temperature, etc. 
These literals require different handling compared to entities. 
One of the issues is that the set of literals is infinite, rather than taken from a finite set. 
This aspect is important both when using these literal values as part of the query, part of the values bound to variables, and when the answer of the query is a literal value.
Overall, one would need to find out how to combine the learning of usable representations of these literal values (i.e., learn a good encoder). 
% COMBINING WITH LMs
One possible direction on how to deal with textual literals in the query is to get initial representations from a language model. 
Given the success these large models have shown in multiple downstream tasks, we can expect that bringing in this knowledge could be beneficial for AQA. 

% USING THEM IRL & EXPLAINABILITY
In terms of explainability, AQA actually has great potential. It is already evident from works such as CQD (explanations via intermediate value predictions) and MPQE and StarQE (explanations by visualising the variable node predictions) that AQA can be explainable with regard to what nodes in the graph seem relevant for the answer.
This partial success, could present itself as a great motivator in adopting this technology for real-life applications. 
An example could be a purely neural QA system that can provide explanations for the answers provided to the user. 
There is, however, no current work which would show why the system decided that these nodes were relevant in the first place. 
For some systems, this goes back to a limited explainability of the underlying link predictor, while in other systems the explaination would be all other queries the system was trianed with, which is so much information that it is hard to distill a humanly understandable explanation.

\noindent Overall, the research on this topic is still very young, but several promising results have already been achieved. 
We foresee much more investigation, not only in new approaches but also in deeper theoretical insight as well as usage of these systems in practical applications.

\bibliographystyle{tfnlm}
\bibliography{reference}

\newpage

\section{Datasets used in the included papers}
\begin{table}[h]
	\centering
	\caption{Overview of used datasets in the investigated papers.}
	\label{tab:query:datasets}
	\begin{tabular}{lccccccccccccc}
		\toprule
		&
		\rotatebox{90}{Bio~\cite{hamilton2018embedding}} &
		\rotatebox{90}{Reddit~\cite{hamilton2018embedding}} &
		\rotatebox{90}{FB15k~\cite{ren2020query2box}} &
		\rotatebox{90}{FB15k-237~\cite{ren2020query2box}} &
		\rotatebox{90}{FB15k-237-CQ~\cite{biqe}} &
		\rotatebox{90}{FB15k-237-PATHS~\cite{biqe}} &
		\rotatebox{90}{NELL995~\cite{ren2020query2box}} &
		\rotatebox{90}{NELL995-CQ~\cite{biqe}} &
		\rotatebox{90}{NELL995-PATHS~\cite{biqe}} &
		\rotatebox{90}{WD50K-QE~\cite{alivanistos2022hyper}} &
		\rotatebox{90}{AIFB~\cite{daza2020message}} &
		\rotatebox{90}{MUTAG~\cite{daza2020message}} &
		\rotatebox{90}{AM~\cite{daza2020message}}
		\\
		\midrule
		AQRE~\cite{wang2018towards} & ? \\
		GQE~\cite{hamilton2018embedding} & \checkmark & \checkmark & - & - & - & - & - & - & - & - & - & - & - \\
		MPQE~\cite{daza2020message} & \checkmark & - & - & - & - & - & - & - & - & - & \checkmark & \checkmark & \checkmark \\
		BiQE~\cite{biqe} & \checkmark & - & - & - & \checkmark & \checkmark & - & \checkmark & \checkmark & - & - & - & - \\
		BetaE~\cite{ren2020beta} & - & - & \checkmark & \checkmark & - & - & \checkmark & - & - & - & - & - & - \\
		Q2B~\cite{ren2020query2box} & - & - & \checkmark & \checkmark & - & - & \checkmark & - & - & - & - & - & - \\
		CQD~\cite{arakelyan2021complex} & - & - & \checkmark & \checkmark & - & - & \checkmark & - & - & - & - & - & - \\
		StarQE~\cite{alivanistos2022hyper} & - & - & - & - & - & - & - & - & - & \checkmark & - & - & - \\
		\bottomrule
	\end{tabular}
\end{table}

\end{document}